# Automatic Vehicle Checking Agent (VCA)

Muhammad Zaheer Aslam[1]  Dr. Shakeel Ahmad [2] Dr. Bashir Ahmad[3]
Zafar Abbas[4]

1. Student of MSCS, Gomal University DlKhan.Pakistan
*zaheer.aslam@yahoo.com

2. Director (IT), Gomal University Dlkhan,Pakistan

3. Director /Associate Professor, Gomal University Dlkhan,Pakistan

4. Student of MSCS, Gomal University Dlkhan,Pakistan

**Abstract**

A definition of intelligence is given in terms of performance that can be quantitatively measured. In this study, we have presented a conceptual model of Intelligent Agent System for Automatic Vehicle Checking Agent (VCA). To achieve this goal, we have introduced several kinds of agents that exhibit intelligent features. These are the Management agent, internal agent, External Agent, Watcher agent and Report agent. Metrics and measurements are suggested for evaluating the performance of Automatic Vehicle Checking Agent (VCA). Calibrate data and test facilities are suggested to facilitate the development of intelligent systems.

**1. Introduction**

As the time passes, there is need of intelligent agent system in place of the human agents. This is because the automatic agents can do the same job in short time, more accurately and more feasible in terms of cost. The agent developed should be intelligent and user-friendly. We have presented a conceptual Automatic Vehicle Checking Agent (VCA) system that can act in the place of a human being. An agent is an information processing. We have assumed that this is a general introductory statement program/model that can be applied to many auto vehicle organizations for producing accurate and fast results. Intelligent agents can be classified into several different categories [1]. Intelligent agents can be non-cooperative and cooperative intelligent agents, depending on their ability to cooperate with each other for the execution of their tasks. The *second* category is referred to as rational intelligent agents and comprises agents that are utilitarian in an economic sense. They act and collaborate to maximize their profit and can be applied to automated trading and electronic commerce. The third class of intelligent agents comprises adaptive intelligent agents that are able to adapt themselves to any type of situation e.g they can be applied to learning personal assistants on the Web. Fourth category which are called mobile intelligent agents are a particular category of agents, which can travel autonomously through the Internet, and can be applied to such tasks as dynamic load balancing among information servers and reducing the volume of data transfers.

The field of intelligent agents has seen rapid growth over the last decade and such agents now constitute powerful tools that are utilized in most industrial applications. Some of the intelligent applications are intelligent user interfaces [2, 1], autonomous agents [3,4], vision systems [5], knowledge discovery and data mining [6], information retrieval [7, 8], electronic commerce [9], personal assistants used on the web [10].

An intelligent agent is generally considered to be an autonomous system that can obtain synergy effects by combining a practical user interface, on the one hand, and an intelligent system based on Artificial Intelligence on the other hand. Intelligent agents are providing a user interface. An Intelligent agent can do the following i.e. it can establish efficient connections between agents, it can also perform job distribution between agents, and the handling of conflicts and errors between agents have not yet been solved.

In this paper, we have investigated a model and design for an intelligent agent system, which helps the user in a user friendly fashion in the auto vehicle industry.

The paper is organized as follows: section 2 describes our Automatic Vehicle Checking Agent System (IAS), section 3 presents the conceptual model of our intelligent agent system i.e. Automatic Vehicle Checking Agent (VCA), section 4 describes the working algorithem of VCA, section 5 presents applications of VCA., section 6 concludes the paper and at the end, section 7 gives the references

## 2. Automatic Vehicle Checking Agent (Vca)

The intelligent agent system described in this paper, referred to as Automatic Vehicle Checking Agent (VCA), consists of 5 agents, Management agent, internal agent, External Agent, watcher agent and report agent symbolized in fig.1.
Before showing the design/conceptual model of VCA we study some requirements of VCA. On which the overall function of VCA will depend.

**Requirement of Automatic Vehicle checking Agent (VCA)**

The features of intelligence required by Automatic Vehicle Checking Agent (VCA) depends on many factors

i. It depends where it is expected to operate.
ii. Different sensors are used which should be available.
iii. It also detects and correct the problem occurred in vehicle.
iv. How it is controlled?
v. What are the costs, risks and benefits?
vi. What skills and ability are required?
vii. What kinds of roads are suitable?

## 3. Design Of Automatic Vehicle Checking Agent (Vca)

The conceptual Model of the Automatic Vehicle Checking Agent (VCA) is depicted in Fig.1. the user interacts with management agent and interact with different agents i.e. Internal Agent, External Agent, Watcher Agent and Report agent and each agent is activate with an appropriate messages. All the agents work together with collaboration and coordination. Each agent has its own assigned role. This is explained below.

### 3.1. Management Agent

The management agent operates in a manner designed to be very friendly to the user. It maintains the agent list, remembers the role of each agent and controls all of the agents.
When the user first addresses the system, it explains the necessary operating procedures. It allows the user's personal profile to be input and stores it in the personal profile database. The interactions between the User and the Management Agent are illustrated in Fig. This agent has overall control over the other agents. If it receives a message from another agent, it selects a suitable agent for the message and activates this agent.

### 3.2. Internal Agent

This agent operates when management agent sends a message for internal checking of vehicle. It checks petrol, Spark plugs, lights, fuel tank problem. Battery problem brake problem and this agent is invoked it send a suitable message to the management agent via sensors because different types are sensors are also used.

### 3.3. External Agent

This agent operates when management agent send a message for external checking of vehicle It checks Tyres, Lights, Door, Speed control. Reversing of vehicle etc speed and this agent is invoked it send a suitable message to the management agent via sensors.

### 3.4. Watcher Agent

This agent watches for messages between agents and keeps track of which agent is currently activated. If there is a conflict between agents or if events do not occur in the right order, it sends a warning message. If a fatal error occurs, it terminates the operation of the currently activated agent and returns to the previous step.

**3.5. Report Agent**
Report agent delivers the final results to the user in a user friendly way by which user can perform his duties.

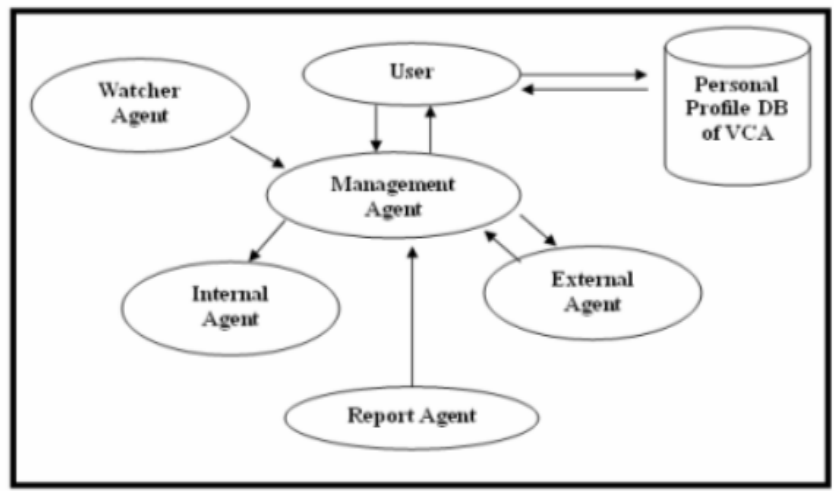

Fig.1  Proposed Model of VCA

**4. Working Of Vca**

**4.1. Pseudo code for Internal Agent**
The main steps of Pseudo Code for Internal Agent are given in the following box

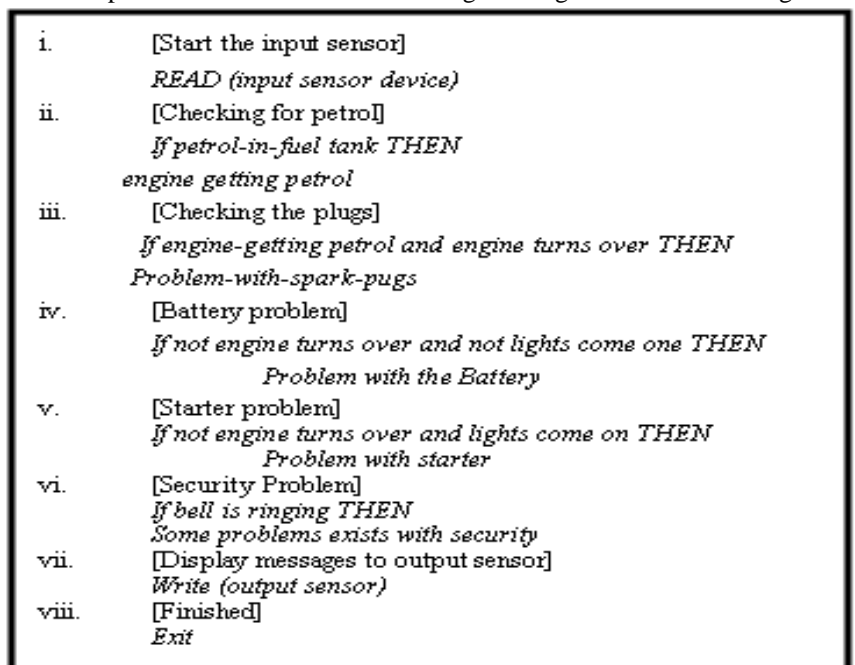

Fig. II Pseudo code of VCA

**Checking by Internal Agent**
  i. If engine getting petrol and engine
        Turns over then problem is with the
              Spark plugs.
 ii. If engine not turn over and lights not come on then problem is with the battery.
iii. If engine not turn over and lights come on then problem is with the starter.

    iv. If petrol not in the fuel tank then Petrol finished
    v. If bell ringing then some one has stolen the vehicle.

### 4.2. Checking by External Agent
    i. It Checks the speed of Vehicle
    ii. Reversing and forwarding of vehicle i.e. forward and back word.
    iii. Damaging or leakage of oil.

### 5. Applications Related To Automatic Vehicle Checking Agent (Vca)

Automatic Vehicle Checking Agent is used to check every vehicle in any auto vehicle organization. The history of every vehicle is also stored in the device which gives all the detail of vehicle and each and every parts of the vehicle. There are many problems in any auto vehicle organization about different parts of an auto vehicle checking i.e. petrol problem, light problem, battery problem, lock problem, engine problem etc Moreover in this automatic vehicle checking agent different sensors are used and each is adjusted in the vehicle to perform its work as an agent and the user is directly attached to a data base of the vehicle as well as the Management agent. When any problem exists then the user activate the management agent then this agent checks that it is an internal problem or external problem, then it activate the agent via sensors which are adjusted in the vehicle for each and every part of vehicle, then an appropriate message is display through watcher agent i.e. Problem with engine, petrol has finished, battery is not working properly, speed of vehicle, damaged in the vehicle etc, after displaying suitable messages through report agent then the user of the vehicle can take an appropriate action.

### 6. Conclusion

In this paper, we have described the modeling and working functions of Automatic Vehicle Checking Agent (VCA) using different kinds of agents, each of which exhibits intelligent features. In future we need to develop a detail description of each agent.

**Dr. Bashir Ahmad** has done PhD in computer Science from Gomal University Dera Ismail Khan Pakistan recently. He is currently the Head/Director of Intiuitiute of Information Technology Gomal University Dera Ismail Khan Pakistan.

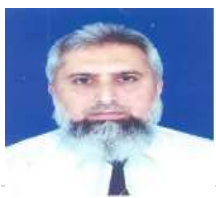

**Dr. Shakeel Ahmad** received his B.Sc. with distinction from Gomal University, Pakistan (1986) and M.Sc. (Computer Science) from Qauid-e-Azam University, Pakistan (1990). He served for 10 years as a lecturer in Institute of Computing and Information Technology (ICIT), Gomal University Pakistan. Now he is serving as an Assistant Professor in ICIT, Gomal University Pakistan since 2001. He is among a senior faculty member of ICIT.

Mr. Shakeel Ahmad received his PhD degree (2007) in Performance Analysis of Finite Capacity Queue under Complex Buffer Management Scheme. Mr. Shakeel's research has mainly focused on developing cost effective analytical models for measuring the performance of complex queueing networks with finite capacities. His research interest includes Performance modelling, Optimization of congestion control techniques, Software refactoring, Network security, Routing protocols and Electronic learning. He has produced many publications in Journal of international repute and also presented papers in International conferences.

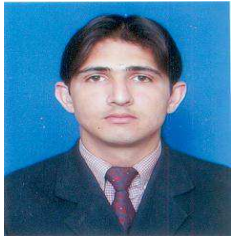

**Mr. Muhammad Zaheer Aslam** has done BS in computer Science from Govt: Degree College NO.1 Dera Ismail Khan affiliated with Gomal University Dear Ismail Khan Pakistan. He has first division throughout his academic carrier. He has done his research on Mobile Adhoc Net (MANET). Currently, I am doing MSCS from gomal University Dera Ismail Khan Pakistan.

**Mr. Zaffar Abbas** has done MSc in computer Science from University of Peshawar, Pakistan. Currently, I am doing MSCS from gomal University Dera Ismail Khan Pakistan.